\documentclass[letterpaper, 10 pt, conference]{ieeeconf}

\usepackage{graphicx} 

\usepackage{subcaption}
\usepackage{colonequals}
\usepackage{amsmath,amsfonts,amssymb}
\usepackage{float}
\usepackage{amsfonts}
 \usepackage{booktabs}
 \usepackage{multirow}
 \usepackage{enumerate}

\DeclareMathOperator*{\argmin}{arg\,min}

\usepackage[ruled]{algorithm2e}
\makeatletter
\newcommand{\algorithmstyle}[1]{\renewcommand{\algocf@style}{#1}}
\makeatother

\usepackage{xcolor}
\usepackage{ulem}

\IEEEoverridecommandlockouts                              

\title{\LARGE \bf
Long range teleoperation for fine manipulation tasks \\under time-delay network conditions
}

\author{Jun Jin$^{1}$, Laura Petrich$^{1}$, Shida He$^{1}$, Masood Dehghan$^{1}$, Martin Jagersand$^{1}$
\thanks{$^{1}$Authors are with Department of Computing Science,
        University of Alberta, Edmonton AB., Canada, T6G 2E1
        {\tt\small jjin5, llorrain, shida3, masood1, jag@ualbertra.ca}}%
}

\begin{document}

\maketitle
\thispagestyle{empty}
\pagestyle{empty}

\begin{abstract}
We present a coarse-to-fine approach based semi-autonomous teleoperation system using vision guidance. The system is optimized for long range teleoperation tasks under time-delay network conditions and does not  require prior knowledge of the remote scene. Our system initializes with a self exploration behavior that senses the remote surroundings  through a freely mounted eye-in-hand web cam. The self exploration stage estimates hand-eye calibration~\cite{horaud1995hand} and provides a telepresence~\cite{sheridan1995teleoperation} interface  via real-time 3D geometric reconstruction. The human operator is able to specify a visual task through the interface and a coarse-to-fine controller guides the remote robot enabling  our system to work in high latency networks. Large motions are guided by coarse 3D estimation, whereas fine motions use image cues (IBVS~\cite{hutchinson1996tutorial}). Network data transmission cost is minimized by sending only sparse points and a final image to the human side. Experiments from Singapore to Canada on multiple tasks were conducted to show our system's capability to work in long range teleoperation tasks.
\end{abstract}

\section{INTRODUCTION}
Long range teleoperation for fine manipulation tasks (e.g., remote facility maintenance, offshore oil rig operations, and space missions) under unknown environments often have limited network conditions. There are two major challenges: (i) For the purpose of situation awareness, traditional telepresence methods~\cite{sheridan1995teleoperation,beck2013immersive}, are no longer practical via low bandwidth network communications as they require real time visual feedback and a fine 3D scene model prior or sound and tactile feedback. (ii) For the purpose of remote control, a fully human supervised system is difficult to use since the response time increases under high latency network conditions. We propose a coarse-to-fine approach to address these challenges.

A coarse-to-fine manipulation approach is commonly used in human eye-hand-coordination when the exact target location or 3D model of the surroundings is unknown. This approach can easily plan the large motion part using a coarse geometric estimation of the scene, and then, if needed, performs fine motions to precisely hit the target using visual texture feedback. Inspired by the same idea, we propose a coarse-to-fine teleoperation system which guides large motions based on a coarse 3D geometric reconstruction and fine motions via visual servoing.

\begin{figure}
	\begin{center}
		\includegraphics[width=0.48\textwidth]{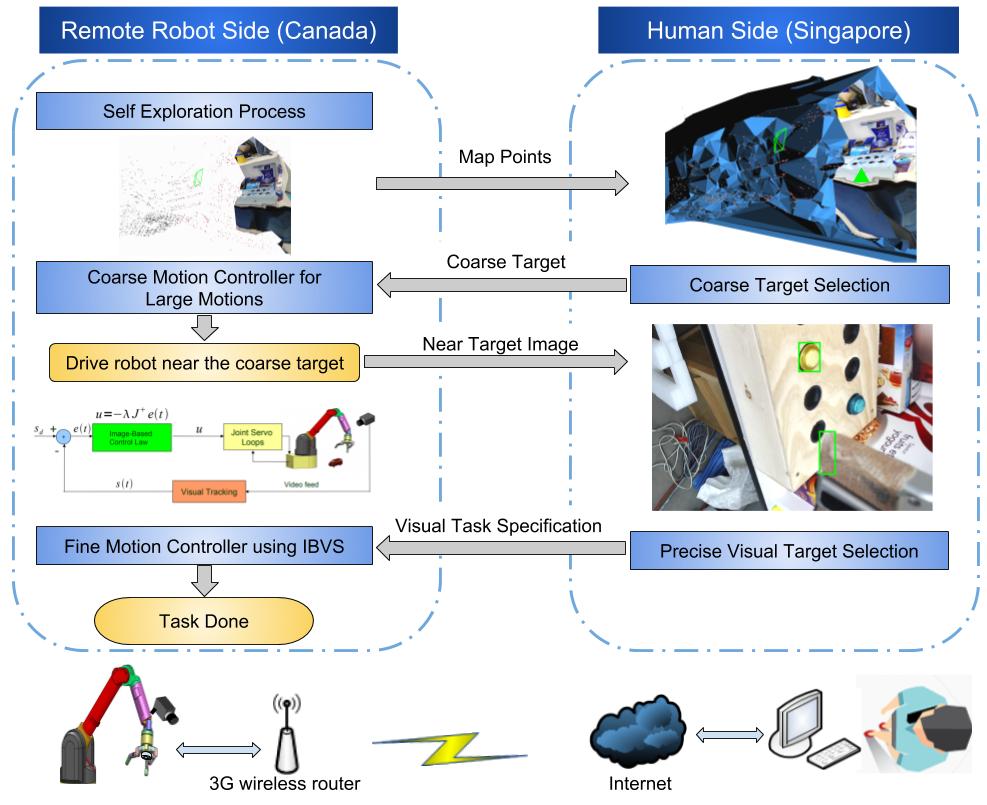} 
		\caption{Design overview: The remote robot begins with a self exploration process which drives an uncalibrated web cam and send back map points. cam poses to human operator. The returned points are used to build a telepresence interface for task specification. Then the robot side coarse-to-fine autonomous controller guides the robot to fulfill the task.}
		\label{fig:design_overview} 
	\end{center}
\end{figure}

Our proposed system starts with a self exploration process to sense the remote surroundings by randomly driving a freely mounted eye-in-hand web cam through the robots workspace. A visual odometry method~\cite{mur2017orb} is used to generate map points throughout this process. After self exploration, hand-eye calibration~\cite{horaud1995hand} is estimated using correspondent map points and a coarse telepresence model is obtained by real-time 3D reconstruction. The human operator then specifies a   target by clicking on the model, which the system will use to generate large motion commands that drive the robot towards the target. Once confirmation of the coarse movement is received, an image of the remote scene is sent to the telepresence interface for further fine task specification  ~\cite{gridseth2013bringing}. Image-Based Visual Servoing (IBVS)~\cite{chaumette2006visual} control is then used to autonomously guide the robot to the desired position. Our major contributions are:
\begin{figure*}
	\centering
	\includegraphics[width=1.0\textwidth]{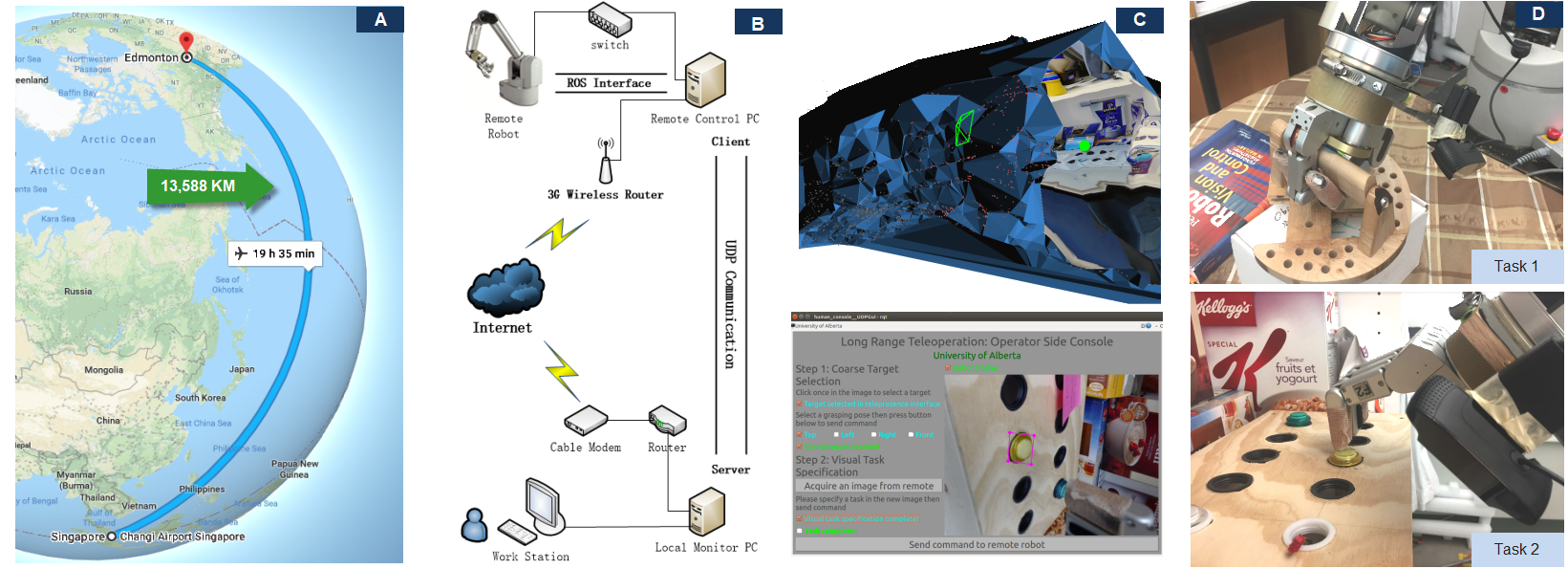}
	\caption{Experiment overview: \textbf{A}. Location of the remote robot and human operator; \textbf{B}. Network structure used for data communication; \textbf{C}. Interface for human operator. Up: Selecting a coarse target. Down: Fine manipulation task specification~\cite{gridseth2013bringing}; \textbf{D}. Two tasks are designed: 1) hold handler. 2) press button. }
	\label{fig:net_design}
\end{figure*}

\begin{itemize}
	\item We propose a long range teleoperation system that works in time-delay network conditions. Capability of our system is demonstrated through experiments using a wireless 3G router to connect between a remote robot and human operator.
	\item Our proposed system includes a coarse-to-fine autonomous control which can work in high latency networks. Errors coming from the coarse 3D model and hand-eye calibration are further compensated in the fine manipulation stage using image cues.
	\item Our proposed system provides telepresence under low bandwidth network by constructing the 3D geometric model of the remote scene.
	\item Our 3D reconstruction process also removes the common constraint in task specification, that the target must be in camera's field of view, by exploiting the coarse 3D model in our visual interface.
	\item The cumbersome camera-robot calibration is automatically calculated in our self exploration stage.
\end{itemize}

\section{Method}
\subsection{Preliminaries}
\begin{figure}
	\begin{center}
		\includegraphics[width=0.3\textwidth]{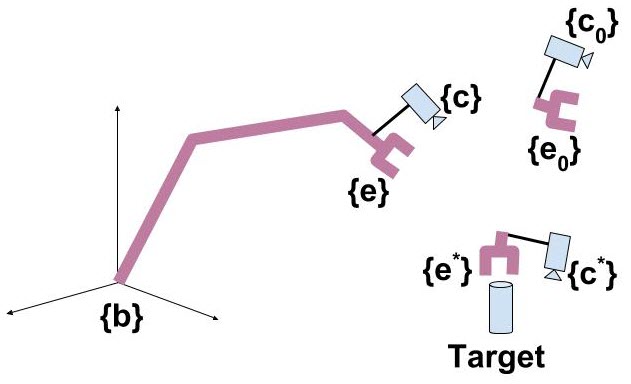} 
		\caption{Coordinate Frames Definition}
		\label{fig:coor_frames}
	\end{center}
\end{figure}

We begin by defining coordinate frames (Fig. \ref{fig:coor_frames}). Our notation style follows the same definition in \cite{corke2017robotics}. Symbols that will be used in further derivation are listed below: 
\begin{table}[H]
\label{table:coordinateFrame}
\begin{center}
\begin{tabular}{@{}cl@{}}
\toprule
\multicolumn{1}{l}{Notations} & Descriptions                          \\ \midrule
$\{b\}$                            & robot's base frame.                   \\
$\{e\}$                            & current end effector's frame.         \\
$\{c\}$                            & current camera's frame.               \\
$\{e_{i}\}$                        & end effector's frame at time i.         \\
$\{c_{i}\}$                        & camera's frame at time i.               \\
$\{e_{0}\}$                        & initial end effector's frame.         \\
$\{c_{0}\}$                        & corresponding initial camera's frame. \\
$\{e^{*}\}$                        & desired end effector's frame.         \\
$\{c^{*}\}$                        & corresponding desired camera's frame. \\
${}^{c_{0}} \xi _{c_{i}}$          & spatial transformation from $\{c_{i}\}$ to $\{c_{0}\}$.\\
${}^{b} \xi _{e_{i}}$             & spatial transformation from $\{e_{i}\}$ to $\{b\}$.\\
${}^{b} \xi _{c_{0}}$        &  spatial transformation from $\{c_{0}\}$ to $\{b\}$.\\
${}^{e} \xi _{c}$              & spatial transformation from $\{c\}$ to $\{e\}$. \\ 
${}^{b}t_{c\rightarrow e}$  & translation from origin of $\{c\}$ to $\{e\}$ defined in $\{b\}$\\
${}^{b}t_{c_{i}\rightarrow e_{i}}$  & translation from origin of $\{c_{i}\}$ to $\{e_{i}\}$ defined in $\{b\}$\\
\bottomrule
\end{tabular}
\end{center}
\end{table}

It's worth noting that: a) the end effector's frame and the camera's frame have a fixed correspondence relationship since they form a rigid body; b) The initial camera's frame $\{c_{0}\}$ is the world coordinate frame used to represent map points, camera poses, and reconstructed 3D model in the self exploration stage.

\subsection{Remote Robot Side: Self Exploration Process} \label{chap:initialization}
This autonomous process drives a freely mounted webcam following a sphere-like routine to explore remote surroundings. It is used to estimate an unknown remote scene, as well as hand-eye calibration using visual odometry.

\subsubsection{Visual Odometry}
In this paper, we use ORB-SLAM\cite{mur2017orb} to generate sparse 3D map points and estimate camera pose. ORB-SLAM is a feature-based method which relies on rich image textures and scene overlaps. Based on the application needs, direct visual odometry methods, such as LSD SLAM~\cite{engel2014lsd} or SVO~\cite{forster2017svo}, may also be used as they generate dense map points which  work well for telepresence visual rendering but generally perform poorly for network data transmission. The visual odometry module incrementally outputs map points ${}^{c_{0}} P^{i}$ defined in Frame \{$c_{0}$\}, as well as continuously estimates the cameras pose.

\subsection{Solving Hand-Eye Calibration}
 Along with the camera pose estimation in the visual odometry module, the corresponding end effector pose is also recorded. Therefore, we are able to construct a dataset with camera poses and end effector poses: $\mathcal{D}=\{({}^{c_{0}} \xi _{c_{i}}, {}^{b} \xi _{e_{i}})\}$, where $i=1,...,n$ defines key frame time steps. By decomposing a SE(3) transformation into a translation part and a rotation part, dataset $\mathcal{D}$ can be further represented as two pairs of correspondence: $\mathcal{D}=\{({}^{c_{0}} t _{c_{i}}, {}^{b} t _{e_{i}}), (^{c_{0}} R _{c_{i}}, {}^{b} R _{e_{i}})\}$. Our \textbf{goal} is to estimate {two spatial transformations}: ${}^{b} \xi _{c_{0}}$ and ${}^{e} \xi _{c}$.

The camera and end effector form a rigid body, namely ${}^{c}R_{e}$ and ${}^{c}t_{ e}$ are fixed for all time steps $i$. Let's denote ${}^{c}t_{ e}$ as  $t_{0}$.
\subsubsection{From initial camera frame to robots base frame ${}^{b} \xi _{c_{0}}$} \label{chap: transform_cam_robot}
Assuming the scale matrix that converts camera frame units to physical world units be denoted as D = diag($\alpha_{x}$, $\alpha_{y}$, $\alpha_{z}$), we have:
\begin{equation}
\label{eq:trans_relationship}
{}^{b}t_{e_{i}}={}^{b}R_{c_{0}} D({}^{c_{0}}t_{c_{i}} +{}^{c_{0}}R_{c_{i}} t_{0}) + {}^{b}t_{c_{0}}
\end{equation} 
where $t_{0}= {}^{c_{i}}t_{e_{i}}$ and ${}^{b}t_{c_{0}}$ does not depend on our parameters ${}^{b}R_{c_{0}}$ and $D$. The objective function is:
\begin{align}
\argmin_{D, {}^{b}R_{c_{0}}, t_{0}} & { \sum_{i=1}^{n}\rVert 	{}^{b}t_{e_{i}} - {}^{b}R_{c_{0}} D({}^{c_{0}}t_{c_{i}} +{}^{c_{0}}R_{c_{i}}t_{0}) \rVert_{2}^{2} } \\
& \text{subject to } {}^{b}R_{c_{0}} \in SO(3). \nonumber
\end{align} 
This forms a relaxed Orthogonal Procrustes Problem \cite{everson1998orthogonal}). We use a tandem algorithm described in \cite{everson1998orthogonal} to solve $D$, ${}^{b}R_{c_{0}}$ and $t_{0}$ by iteratively optimizing one while fixing others. 

In our experiments, we simply assume $t_{0}=\mathbf{0}$ as the camera and end effector are situated close to each other, since this error will be further compensated in the fine motion control part using image textures.

Subsequently, ${}^{b}t_{c_{0}}$ can be calculated using an average estimation from all samples. Now we have estimated $D$, ${}^{b}R_{c_{0}}$ and ${}^{b}t_{c_{0}}$. The transformation from $\{c_{0}\}$ to $\{b\}$, i.e., ${}^{b} \xi _{c_{0}}$ is solved.

\begin{figure}
	\centering
	\includegraphics[width=0.3\textwidth]{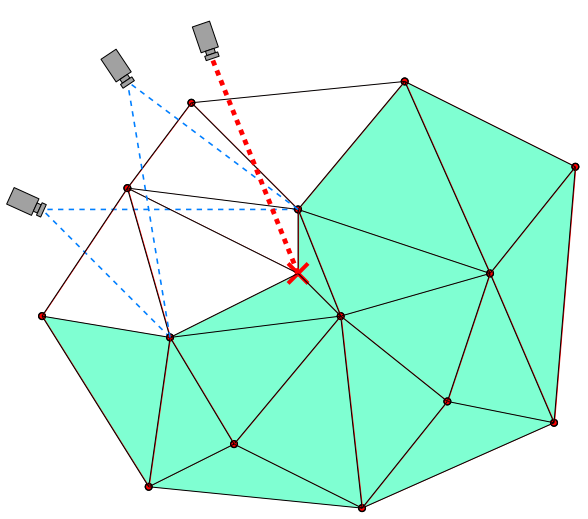}
	\caption{The CARV~\cite{lovi2011incremental} algorithm is based on free-space constraint and has the ability to incrementally refine the 3D reconstruction result in real-time. Basic tetrahedron cells are built by triangulation and refined as new points are added (red cross).}
	\label{fig:CARV_System}
\end{figure}
  
\subsubsection{From camera frame to end effector frame ${}^{e} \xi _{c}$} \label{chap:hand2eye}
By using orientation correspondences (${}^{c_{0}} R _{c_{i}}$, ${}^{b} R _{e_{i}}$), we have:
 \begin{equation}
\label{eq:orientation}
{}^{b} R _{e_{i}} =  {}^{b}R_{c_{0}}{}^{c_{0}}R_{c_{i}} {}^{c} R _{e}
\end{equation}
where ${}^{c} R _{e}={}^{c_{i}} R _{e_{i}}$. Now the cost function is:
\begin{align}
\argmin_{{}^{c} R _{e}} & {\sum_{i=1}^{n}\rVert {}^{b} R _{e_{i}} -  {}^{b}R_{c_{0}}{}^{c_{0}}R_{c_{i}} {}^{c} R _{e} \rVert_{2}^{2} } \\
& \text{subject to} {}^{e} R _{c} \in SO(3). \nonumber
\end{align} 

This is a regular Orthogonal Procrustes Problem. A general solution \cite{schonemann1966generalized} based on SVD can solve ${}^{c} R _{e}$. Also, $t_{0}={}^{c}t_{e}$ is solved previously, now the transformation from $\{e\}$ to $\{c\}$, i.e., ${}^{e} \xi _{c}$ is obtained.

\subsection{Human Side: Visual Telepresence Interface}
\subsubsection{Real-time 3D reconstruction}
Given a set a map points $\{{}^{c_{0}} P^{i}\}$ that are transmitted to human side, real-time 3D geometric reconstruction methods~\cite{lovi2011incremental}, \cite{niessner2013real}, \cite{he2018incremental} can be used to build a visual telepresence interface. We use CARV~\cite{lovi2011incremental} in this paper (as shown in Fig. \ref{fig:CARV_System}), other more sophisticated methods~\cite{petit2010multicamera} can also be used to create better rendering effects. The reconstructed 3D geometric model generated in CARV is defined in frame $\{c_{0}\}$.

\subsubsection{Task specification}
\begin{figure}
	\centering
	\includegraphics[width=0.48\textwidth]{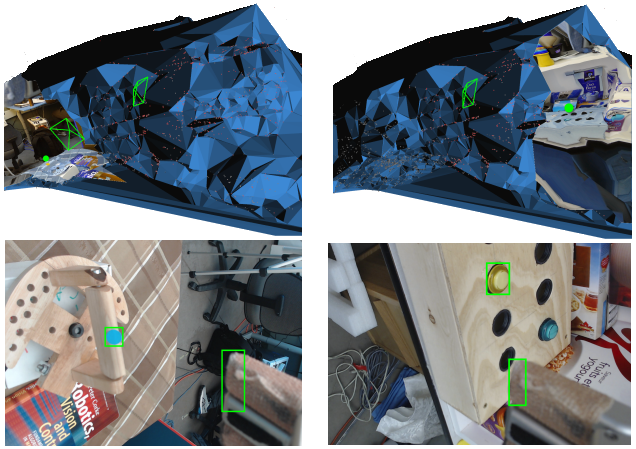}
	\caption{Visual telepresence interface for task specification. \textbf{Left Column}: Hold handler task; \textbf{Right Column}: Press button task; \textbf{Up Row}: Coarse target selection; \textbf{Down Row}: Precise task specification.}
	\label{fig:task_specification}
\end{figure}

\begin{figure}
	\centering
	\includegraphics[width=0.4\textwidth]{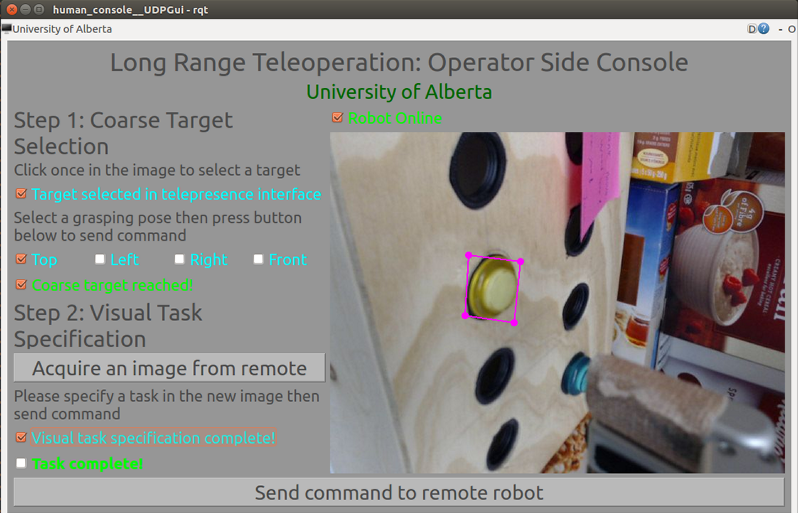}
	\caption{Human side console interface. It allows user to select a preferred end effector orientation. It also monitors remote task execution status and pops up an image for precise task specification as shown in Fig. \ref{fig:task_specification}}
	\label{fig:select_orientation}
\end{figure}

As shown in Fig. \ref{fig:task_specification}, the operator is required to select a target on the rendered 3D model with a desired end effector orientation. This is called coarse target selection. A console interface is designed (Fig. \ref{fig:select_orientation}) for the user to select among 4 preset orientations. The task command is then send back to the remote robot side as:
\begin{itemize}
	\item ${}^{c_{0}}{P^{*}}$: Position of target point \textbf{$P^{*}$} defined in \{$c_{0}$\}.
	\item ${}^{b}R_{e^{*}}$: Desired end effector orientation defined in \{b\}.
\end{itemize}

After receiving the task command, a coarse controller will drive the robot towards the target. Since the camera is also near the target now, an image is sent back to human side for precise task specification. As shown in~\cite{gridseth2013bringing}, we use a similar interface to specify a task using image features. The human operators selection is then sent to the remote robot side, where an IBVS controller is used to fulfill the task.

\subsection{Remote Robot Side: Coarse-to-Fine Controller}
\subsubsection{Coarse motion controller for large motions}
The main purpose of our coarse controller is to bring the target into the camera's field of view by driving the robot towards it, thus fulfilling the large movement part in task execution, which is why a coarse 3D model still works in practice. 

One way to drive large motions is to use an inverse kinematics solver. Since all of the calibration transformations are already complete, it is trivial to get our desired end effector's position ${}^{e}{P^{*}}$ if we substitute ${}^{c_{0}}t_{c_{i}}$ with ${}^{c_{0}}{P^{*}}$ in eq. (1).

However, an inverse kinematics solver may cause large joint movements, which is unsafe in practice. For smooth motion purpose, PBVS~\cite{hutchinson1996tutorial} is used for coarse motion control. Let's denote ${}^{c_{0}}{P^{*}}$ as ${}^{c_{0}}t_{e^{*}}$, since target point $P^{*}$ can represent the desired end effector position. Given the transformation matrix, it's trivial to derive basic equations of PBVS control as:

\begin{equation}
\label{eq:task_c0}
\begin{cases}
{}^{c_{0}}R_{c^{*}} = ({}^{b}R_{c_{}0})^{\intercal}{}^{b}R_{e^{*}}{}^{e^{*}}R_{c^{*}}\\
{}^{c_{0}}t_{c^{*}} = {}^{c_{0}}t_{e^{*}} - {}^{c_{0}}R_{c^{*}} {}^{c^{*}}t_{e^{*}}
\end{cases}
\end{equation}
where ${}^{e^{*}}R_{c^{*}}={}^{e}R_{c}$ and ${}^{c^{*}}t_{e^{*}} = t_{0}$. Now we have the desired camera pose. Given current camera pose (${}^{c_{0}}t_{c}$, ${}^{c_{0}}R_{c}$), we can represent error vector in desired camera's frame $\{c^{*}\}$ as $\mathbf{e}=(\mathbf{e}_{l}, \mathbf{e}_{\omega})$:
\begin{equation}
\label{eq:error}
\begin{cases}
\mathbf{e}_{l} = {}^{c^{*}}R_{c_{0}} ({}^{c_{0}} t _{c} - {}^{c_{0}}t_{c^{*}})\\
\mathbf{e}_{\omega} = ({}^{c_{0}}R_{c^{*}})^{T}\:{}^{c_{0}}R_{c}
\end{cases}
\end{equation}

At last, a simple PD control law can be designed~\cite{hutchinson1996tutorial,kragic2003robust} minimizing the error to zero.

\subsubsection{Fine motion controller using image cues}\label{chap:coarse_controller}
After human operator specifies the task on the send-back image (Fig. \ref{fig:task_specification}), fine motion controller will be activated. 

Image based visual servoing \cite{chaumette2006visual} is used as our fine motion controller. IBVS can both work in calibrated and uncalibrated scenarios. IBVS has the ability to compensate errors in calibration by directly minimizing errors represented in image space. 
\section{Experimental Setup}
\subsection{Network Structure}
Work sites for long range teleoperation tasks (e.g., offshore oil rig operation, remote facility maintenance) often have poor network infrastructures. Considering this practical issue, we use a 3G wireless router which connects the remote robot to the Internet. Thus, a human operator can work in any locations where there is Internet access (as shown in Fig. 2). 

A data communication module based on UDP protocal is designed to transmit map points and images from the remote robot side to the human operator. Task specification data is also sent to robot side for autonomous control purpose. Detailed network structure is shown in Fig. 2.

\begin{figure}[]
	\centering
	\includegraphics[width=0.4\textwidth]{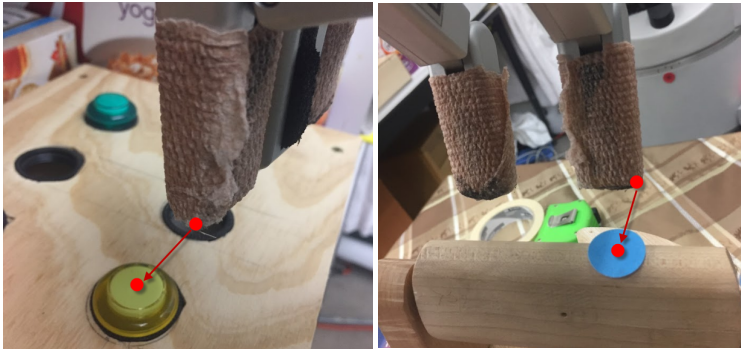}
	\caption{Measurement of the final error. For all the two tasks, error is measured as the distance from the final end effector position (finger tip) to the target. \textbf{Left}: hold handler task; \textbf{Right}: press button task. A threshold of 10mm is used to determine success of a trial. The final grasping motion is hard coded since it's simply a close grasp action.}
	\label{fig:sas}
\end{figure}
\begin{table}[]
\begin{center}
		\caption{Comparison to baseline method. Results show that our method has higher success rate, less task execution time and lower network data consumption. The self exploration (duration: 471s, datagram size: 4.154KB) phase in our method is only required to do once. All following tasks can share the same model thus task duration and datagram size are reduced.}
		\label{my-label}
\begin{tabular}{@{}rcccc@{}}
\toprule
\multicolumn{1}{c}{\multirow{2}{*}{Metrics}} & \multicolumn{2}{c}{Hold handler (10 trials)} & \multicolumn{2}{c}{Press button (10 trials)} \\ \cmidrule(l){2-5} 
\multicolumn{1}{c}{}                         & Baseline        & \textbf{Ours}              & Baseline        & \textbf{Ours}              \\ \midrule
Success rate (\%)                            & 70              & \textbf{100}               & 70              & \textbf{90}                \\
Task duration (s)                            & 1,778.2         & \textbf{471.0+36.0}             & 3,318.2         & \textbf{471.0+125.0}             \\
Data size (KB)                           & 10,812.0        & \textbf{4,154+323.8}       & 20,540.0        & \textbf{4,154+376.5}       \\ \bottomrule
\end{tabular}
\end{center}
\end{table}

\begin{figure*}
	\centering
	\includegraphics[width=1.0\textwidth]{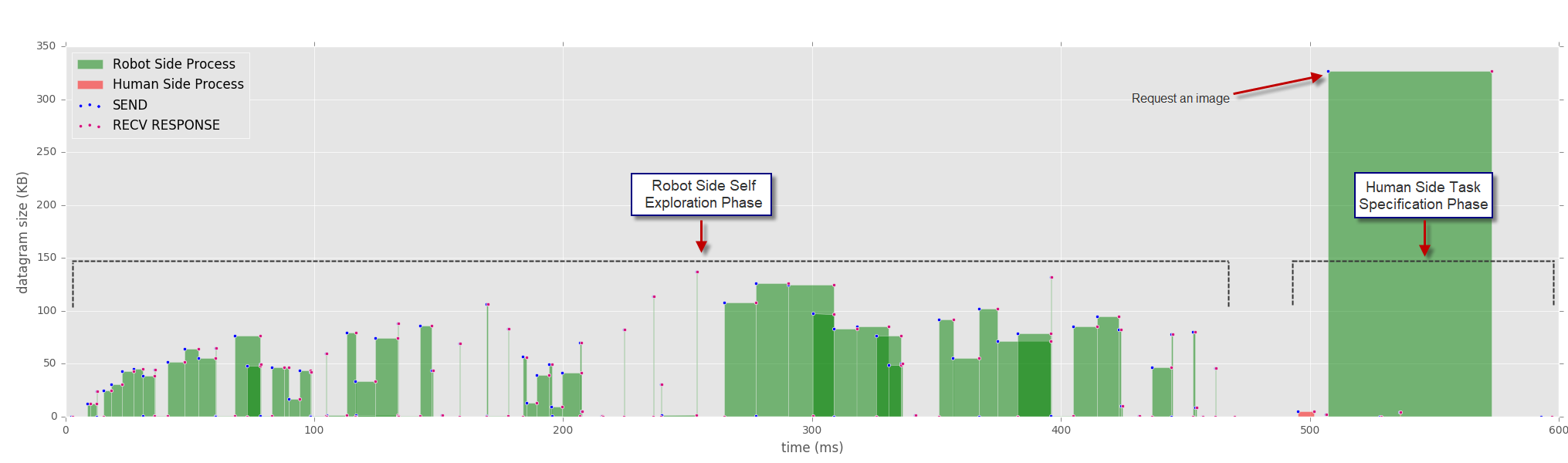}
	\caption{Measurement of datagram transmitted via network on task 2: Press Button. \textbf{x axis}: Timeline (s). \textbf{y axis}: Datagram size(KB). \textbf{Blue dot}: Timestamp of sending data. \textbf{Red dot}: Timestamp of receiving response. \textbf{Green Line}: Time delay measured on robot side (from sending to receiving response). \textbf{Red Line}: Time delay measured on human side (from sending to receiving response). \textbf{The average time delay is 111.4 ms/KB.} The majority of total task duration and network data consumption are in our system's self exploration phase. However, this part is only required to do once at the beginning. Following tasks will be done using the same coarse model.}
	\label{fig:network_measurement}
\end{figure*}
\subsection{Teleoperation Tasks}
Two tasks were carried out in our experiments (as shown in Fig. 2). Both tasks require the remote robot to coarsely estimate its surroundings and return a telepresence model to human operator. Task 1 is a regular manipulation type, while task 2 requires fine manipulation.
\begin{itemize}
	\item Task 1: Holder handle. Robot is required to hold a handle based on human operator's selection.
	\item Task 2: Press button. Robot is required to press a button according to human operator's selection. 
\end{itemize}
\section{Evaluation Results}
\subsection{Comparison to baseline method}
\subsubsection{Baseline method}
Our baseline method is a Cartesian controller which relies on an inverse kinematics solver from MoveIt~\cite{moveit}. Remote images are continuously transmitted to the human side,  requiring on average 35.2s per frame. There are two intrinsic limitations in baseline method compared to ours:
\begin{enumerate}[(a)]
	\item A target must be always in the camera's field of view, otherwise a human operator has no idea where the target may be. This is tedious since the view field of an eye-in-hand camera is relative small.
	\item Task execution is fully controlled by human side, which can be challenging and frustrating for the user in practice because of the large response delay.
\end{enumerate}

In order to compare, we made two assumptions: a) Human operator knows roughly where the target is  to ensure the baseline method is possible to use. b) Task duration is unlimited, unless the human operator chooses to abort early.

\subsubsection{Comparison metrics}
Both the baseline method and our methods include 10 trials of the above mentioned two tasks. We designed 4 metrics for comparison: 
\begin{enumerate}[(a)]
	\item Success rate of trials, which is based on a threshold (10mm) of the final error.
	\item Average time to complete task (min). In failed trials, we set it to 100 min for convenience in calculation.
	\item Average network data consumption (kB), which measures the total size of datagrams transmitted via network.
\end{enumerate}

Evaluation results are shown in Table \ref{my-label}.

\subsection{Measurement of data communication}
For the purpose of analyzing how time-delay network conditions affect the communication between remote robot controller and human operator, we measure two factors in our method: a) datagram size (KB) transmitted via network for all tasks; b) round trip response delays (s) caused by network latency. Since we have the UDP communication interface for both robot side and human side, the above mentioned measurements are taken by logging, sending, and receiving datagrams in both the client and server side. Each datagram will have a short response ending with its datagram ID. So that a round trip delay can be measured without synchronizing the client and server side time. Measurement results are shown in Fig. \ref{fig:network_measurement}. The average time delay in our `press button' task is 111.4 ms/KB.

\section{Related Works}
Long range teleoperation control has lots of work done using Internet connections~\cite{hu2001internet}, \cite{uddin2017long}. While it provides a possibility for convenient operation regardless of location, it typically omits a common limitation in practice: network conditions. Low bandwidth and high latency networks are more common, for example, in tasks such as space missions (e.g., 6 to 44 minutes delay in Martian Rover~\cite{longrange2018})  where an Internet connection is not available. 

Teleoperation under such network conditions is a significant challenge. Khan \cite{khan2010wireless} proposed a wireless network architecture to optimize data transition and control, thus improving network quality. Yokokohji et al.\cite{yokokohji2001bilateral} proposed a PD-based bilateral control method testing on a satellite (ETS\_VII) with round-trip delay 6 seconds. However, addressing both telepresence and control in time-delay network conditions is rare. One possible solution is to use robotic vision. Lovi et al.~\cite{lovi2010predictive} proposed a method using PTAM~\cite{klein2007parallel} and CARV~\cite{lovi2011incremental}  with a telepresence interface and predictive display to generate rendered intermediate image for human control guidance. However, previous images are needed for a projective texture rendering which requires high network consumption and a fully human supervised control causes difficulties in practice.

For the 3D reconstruction, there are other approaches (e.g., multi-camera based method~\cite{petit2010multicamera} and RGB-D data based methods~\cite{niessner2013real}), that can create a  usable telepresence model, however, they typically require high performance network conditions and machines. For robot side semi-autonomous control using vision guidance, Gridseth et al.~\cite{gridseth2013bringing} proposed an interface for task specification based on geometric constraints with error mapping represented in image space. A successive uncalibrated IBVS~\cite{chaumette2006visual} controller is used for robot side autonomy.

\section{Conclusions}
We present a coarse-to-fine approach based teleoperation system taking into consideration the common network limitation in long range telerobotic control: low bandwidth and high latency. The `coarse' part includes a coarse estimation of the remote scene and a controller for large movements. The `fine' part includes an IBVS~\cite{chaumette2006visual} controller for fine motions, which compensates errors in the coarse part by minimizing errors in image space. Only sparse points, an target image and two task commands are transmitted via the network, thus removing the low bandwidth challenge. The robot side autonomous controller removes the high latency challenge. Experiments are designed to showcase our methods ability in such network conditions.

\textbf{Limitations and future work}: We use Orb-SLAM~\cite{mur2017orb} to generate map points. However, this feature based method relies highly on rich image textures. In scenarios with a lack of image textures, direct methods (e.g., LSD SLAM~\cite{engel2014lsd}) can be used. However, for poor illumination scenarios (e.g., space missions), both of these methods will not perform well. Furthermore, a pure coarse 3D model based telepresence may still cause human understanding difficulties. Generating denser map points may help, however, it consumes a larger network cost. The second limitation comes from our fine manipulation controller: IBVS~\cite{chaumette2006visual}, which relies on trackers to estimate error in image space. This is a typical challenge in visual servoing. Several approaches are proposed, e.g., direct visual servoing~\cite{silveira2012direct}, however new challenges arise~\cite{bateux2018going}. Other learning based methods~\cite{jin2018robot}, \cite{levine2016end} could be a future direction. 

We are planning to deploy our system on a mobile manipulator. Combined with a motion planning module, our system can work for tasks requiring both self-navigation and manipulation, enhancing application potential in remote facility maintenance tasks, offshore oil rig operations and space missions.

\addtolength{\textheight}{-12cm}   





\bibliographystyle{IEEEtran}
\bibliography{IEEEabrv,IEEEexample}

\end{document}